\documentclass[pdflatex,sn-mathphys-num]{sn-jnl}

\usepackage{graphicx}%
\usepackage{multirow}%
\usepackage{amsmath,amssymb,amsfonts}%
\usepackage{amsthm}%
\usepackage{mathrsfs}%
\usepackage[title]{appendix}%
\usepackage{xcolor}%
\usepackage{textcomp}%
\usepackage{manyfoot}%
\usepackage{booktabs}%
\usepackage{algorithm}%
\usepackage{algorithmicx}%
\usepackage{algpseudocode}%
\usepackage{listings}%
\usepackage{enumitem}
\usepackage{cleveref}
\usepackage{adjustbox}

\theoremstyle{thmstyleone}%
\theoremstyle{thmstyletwo}%

\theoremstyle{thmstylethree}%

\newcommand{\dataset}{Stanford Sleep Bench}

\begin{document}

\title[Article Title]{{\dataset}: Evaluating Polysomnography Pre-training Methods for Sleep Foundation Models}


\author[1,2,3,*]{\fnm{Magnus Ruud} \sur{Kjaer}}\email{magnusrk@stanford.edu}
\author[4, 5,*]{\fnm{Rahul} \sur{Thapa}}\email{rthapa86@standford.edu}
\author[1]{\fnm{Gauri} \sur{Ganjoo}}
\author[1]{\fnm{Hyatt} \sur{Moore IV}}
\author[3,6]{\fnm{Poul Jørgen} \sur{Jennum}}
\author[7]{\fnm{Brandon} \sur{M. Westover}}
\author[4]{\fnm{James} \sur{Zou}}
\author[1,+]{\fnm{Emmanuel} \sur{Mignot}}
\author[5,+]{\fnm{Bryan} \sur{He}}
\author[2,+]{\fnm{Andreas} \sur{Brink-Kjaer}}\email{andbri@dtu.dk}

\affil[1]{Department of Psychiatry and Behavioral Sciences, Stanford University, Stanford, CA, USA}
\affil[2]{Department of Health Technology, Technical University of Denmark, Kongens Lyngby, Denmark}
\affil[3]{Department of Clinical Neurophysiology, Danish Center for Sleep Medicine, Copenhagen University Hospital – Rigshospitalet, Copenhagen, Denmark} 
\affil[4]{Department of Biomedical Data Science, Stanford University, Stanford, CA, USA}
\affil[5]{Department of Computer Science, Stanford University, Stanford, CA, USA}
\affil[6]{Department of Clinical Medicine, University of Copenhagen, Copenhagen, Denmark}
\affil[7]{Department of Neurology, Beth Israel Deaconess Medical Center, Harvard Medical School, Boston, MA, USA}
\affil[*]{Co-first authors}
\affil[+]{Co-senior authors}


\abstract{Polysomnography (PSG), the gold standard test for sleep analysis, generates vast amounts of multimodal clinical data, presenting an opportunity to leverage self-supervised representation learning (SSRL) for pre-training foundation models to enhance sleep analysis. However, progress in sleep foundation models is hindered by two key limitations: (1) the lack of a shared dataset and benchmark with diverse tasks for training and evaluation, and (2) the absence of a systematic evaluation of SSRL approaches across sleep-related tasks. To address these gaps, we introduce {\dataset}, a large-scale PSG dataset comprising 17,467 recordings totaling over 163,000 hours from a major sleep clinic, including 13 clinical disease prediction tasks alongside canonical sleep-related tasks such as sleep staging, apnea diagnosis, and age estimation. We systematically evaluate SSRL pre-training methods on {\dataset}, assessing downstream performance across four tasks: sleep staging, apnea diagnosis, age estimation, and disease and mortality prediction. Our results show that multiple pretraining methods achieve comparable performance for sleep staging, apnea diagnosis, and age estimation. However, for mortality and disease prediction, contrastive learning significantly outperforms other approaches while also converging faster during pretraining. To facilitate reproducibility and advance sleep research, we will release {\dataset} along with pretrained model weights, training pipelines, and evaluation code.}




\maketitle

%

\section{Introduction}
\label{section_intro}

Sleep plays a pivotal role in maintaining health and well-being, and its disruption has been linked to increased risk for many medical conditions, including neurodegenerative diseases \cite{gottesman2024impact, wunderlin2020role}, metabolic disorders \cite{grandner2012sleep, zhang2023poor, che2021association}, and cardiovascular diseases \cite{pengo2022sleep, grandner2012sleep, ayafor2020association}. 
Polysomnography (PSG) is a critical tool in sleep research and clinical practice. PSG studies are typically performed in sleep clinics and involve full-night recordings of multiple physiological signals, including brain activity signals (BAS), respiratory signals (RESP), electrocardiography (ECG), and electromyography (EMG) \cite{kryger2010principles}.
Reflecting the critical role of sleep research and clinical care, hundreds of thousands of sleep studies are performed annually in the U.S., serving as a cornerstone for diagnosing and managing conditions such as sleep apnea, narcolepsy, and REM sleep behavior disorder \cite{chiao2017trends}.

Clinical use of PSG recordings requires the labeling of sleep stages and apnea events \cite{berry2017aasm, rundo2019polysomnography}.
PSG datasets can also be enriched with medical record data about patients’ health, encompassing psychiatric disorders, neurological conditions, and cardiovascular conditions \cite{quan1997sleep}. With the advent of deep learning, numerous studies have leveraged such data to automate sleep staging \cite{perslev2021u, Stephansen2018NeuralNarcolepsyb,Supratak2020TinySleepNet:EEG, Mousavi2019SleepEEGNet:Approach,Seo2020Intra-andEEG,Phan2021XSleepNet:Staging}, automate diagnosis of sleep apnea \cite{zahid2023msed,kjaer2024abed}, predict age and mortality \cite{brink2022age}, and model other health outcomes such as dementia \cite{ye2023dementia}, stroke \cite{xie2021ischemic}, and heart failure \cite{yan2021objective}. Although these methods have shown great promise, they rely heavily on labeled data, which requires sleep experts to manually annotate the recordings, a time-consuming and labor-intensive task prone to noise and inconsistencies due to differences in labeling procedures, site variability, and subjective interpretations \cite{liu2021large}.


\begin{figure}[t!]
    \centering
    \includegraphics[width=0.99\linewidth]{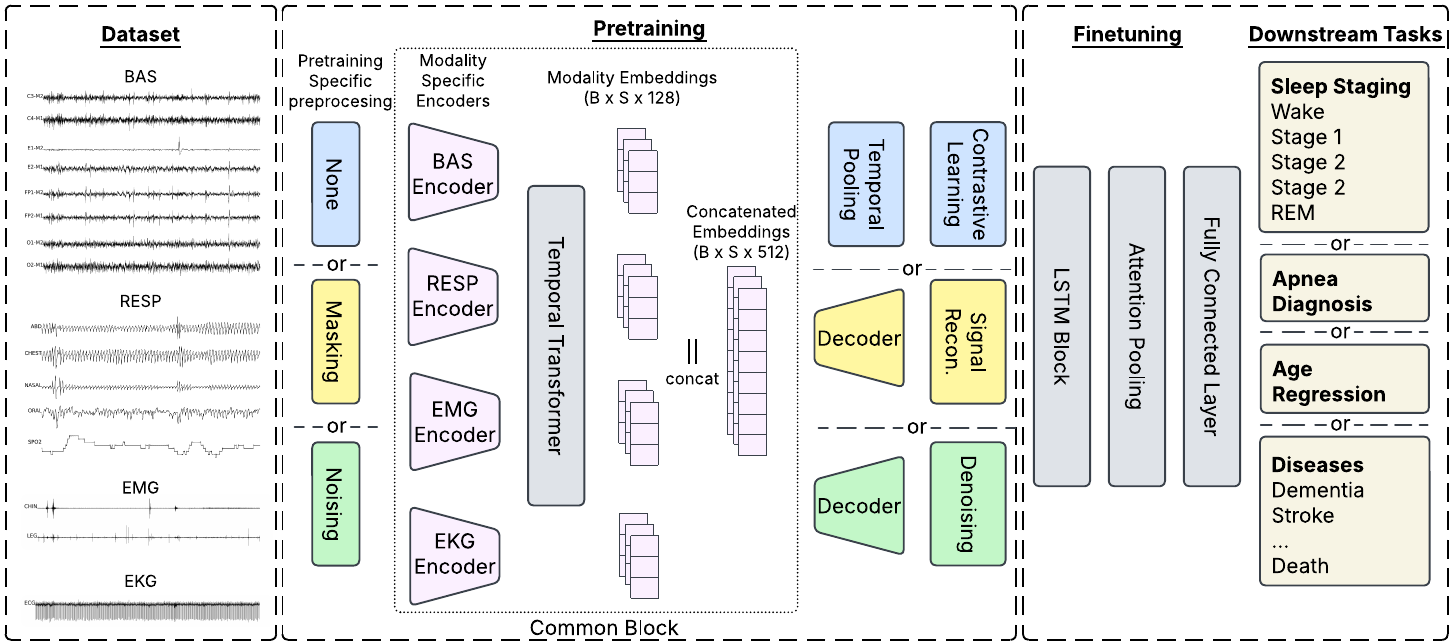}
    \caption{Overview of {\dataset} and the training pipeline for multiple SSRL methods. {\dataset} includes four signal modalities: brain activity (BAS), respiration (RESP), electrocardiogram (EKG), and electromyography (EMG). Each modality is encoded independently using a CNN-based encoder, followed by a temporal transformer. SSRL methods—contrastive learning, signal reconstruction, and denoising—are trained separately, with no shared parameters. After pre-training, an LSTM-based prediction head with attention pooling is added for downstream tasks. Note that while all SSRL methods are shown, they are trained as separate models.}
    \label{fig:overview}
\end{figure}

Foundation models, particularly those leveraging self-supervised representation learning (SSRL), offer a promising alternative to address these challenges \cite{bommasani2021opportunities}. SSRL methods have achieved success by leveraging large-scale unlabeled datasets to learn rich representations. These methods have been applied successfully in biomedicine, including pathology analysis \cite{xu2024whole}, genomic sequence modeling \cite{nguyen2024sequence}, and biosignal processing \cite{abbaspourazad2023large, narayanswamy2024scaling} tasks. In sleep research, preliminary work has explored using foundation models to address tasks such as sleep staging and apnea diagnosis \cite{Thapa2024SleepFM:Signals}. By leveraging multimodal sleep data, foundation models can improve performance on tasks ranging from apnea diagnosis to disease diagnosis. Although PSG data is available from resources such as the National Sleep Research Resource (NSRR), clinical labels such as mortality and diagnoses are typically unavailable, and dataset splits and evaluation metrics are not standardized. Thus, it is difficult to discern the best methods for sleep-related tasks. The effectiveness of different SSRL methods has also not been systematically evaluated for sleep, leaving a critical gap in understanding how to best exploit PSG data for downstream applications. In this work, we make the following \textbf{contributions}:~

\begin{itemize}[leftmargin=.2cm]
    \item \textbf{Dataset and Benchmark:} We introduce {\dataset}, a large-scale PSG dataset and benchmark comprising 17,467 PSG recordings spanning 163,650 hours. {\dataset} includes standardized training, validation, and test splits, supporting 12 clinical disease and mortality prediction tasks alongside canonical sleep-related tasks such as sleep staging, apnea diagnosis, and age estimation. 

    \item \textbf{Systematic Evaluation:} We evaluate a variety of self-supervised representation learning (SSRL) methods applied to PSG data, including masked autoencoders, denoising, and contrastive learning (CL), and assess their performance on four downstream tasks: sleep staging, apnea diagnosis, age estimation, and disease and mortality prediction. Our findings show that while multiple pre-training methods achieve comparable performance for sleep staging (AUROC: $0.802-0.823$), apnea diagnosis (AUROC: $0.818-0.830$), and age estimation (mean absolute error: $6.197-6.666$ years), CL significantly outperforms other approaches for mortality and disease prediction, achieving an average improvement of $4.64\%$ ($0.62\%-7.41\%$).

    \item \textbf{Open-Source Release:} To support sleep research, we provide pre-trained weights, training and evaluation code, and a few-shot benchmark. Together with {\dataset}, these resources promote reproducibility and provide a standardized platform for advancing sleep foundation models. 
\end{itemize}  



%
%

\section{Related Work}
\label{section_related_work}


\noindent\textbf{Machine Learning in Sleep Analysis} \hspace{0.5em}
Machine learning (ML) has played an increasing role in sleep research, particularly in automating canonical sleep scoring tasks such as sleep staging \cite{Stephansen2018NeuralNarcolepsyb, Supratak2020TinySleepNet:EEG, Mousavi2019SleepEEGNet:Approach, Seo2020Intra-andEEG, Phan2021XSleepNet:Staging, perslev2021u} and apnea diagnosis \cite{zahid2023msed, nassi2101automated}. Beyond sleep-specific tasks, ML-based approaches have enabled broader clinical predictions using PSG data, with feature-based models associating sleep disturbances with disease risks, including neurodegenerative diseases and cardiovascular conditions \cite{HjulerAndersen2024ProbabilisticDisorders, Stephansen2018NeuralNarcolepsyb}.

Traditional ML approaches relied on hand-engineered features extracted from polysomnography (PSG) signals, while deep learning models have demonstrated superior performance by directly learning feature representations from raw signals. These deep learning methods have advanced sleep disorder diagnosis \cite{gunter2023svit, brink2022end} and physiological aging assessment \cite{brink2022age}. However, most of these models rely on supervised learning and annotated datasets, limiting their generalizability across different clinical settings.

\noindent\textbf{Foundation Models and Time Series Representation Learning} \hspace{0.5em}
Foundation models have transformed multiple domains by learning general-purpose representations from large-scale unlabeled data, primarily through self-supervised representation learning (SSRL). Common SSRL strategies include masked prediction, denoising autoencoders (DAE), and contrastive learning. Masked prediction tasks, such as masked language modeling in BERT \cite{devlin2018bert} and masked autoencoders for images \cite{he2022masked}, train models to reconstruct missing regions, while denoising autoencoders learn robust representations by removing noise from corrupted inputs \cite{vincent2008extracting, gondara2016medical}. Contrastive learning, as used in MoCo \cite{he2020momentum}, SimCLR \cite{chen2020simple}, and CLIP \cite{radford2021learning}, learns by distinguishing between similar and dissimilar instances.

Recent advances have extended these techniques to time series data, with MOMENT \cite{goswami2024moment} and Moirai \cite{woo2024unified} applying masked prediction for general time series analysis, while TimesFM \cite{das2023decoder} introduces a decoder-only architecture optimized for forecasting. GPT4TS \cite{zhou2023one} adapts large language model architectures for time series modeling. In healthcare, foundation models such as ClinicalTime \cite{liu2024generalized} have demonstrated the effectiveness of self-supervised learning for sparse and irregular clinical time series. Sleep-specific foundation models, including SleepFM \cite{Thapa2024SleepFM:Signals}, leverage multimodal PSG data to enhance sleep staging and apnea diagnosis. However, the development of foundation models for sleep analysis remains limited by the lack of a shared dataset and benchmark for training and evaluation, as well as the absence of a systematic comparison of existing SSRL approaches across sleep-related tasks.

\noindent\textbf{Polysomnography Datasets} \hspace{0.5em}
The National Sleep Research Resource (NSRR) \cite{zhang2018national} is the primary repository for shared PSG datasets. Notable collections include the Multi-Ethnic Study of Atherosclerosis (MESA) \cite{chen2015racial}, which analyzed 2,237 participants aged 45–84 between 2010–2012; the Outcomes of Sleep Disorders in Older Men (MrOS) \cite{blackwell2011associations}, which assessed 3,930 men aged 65+ between 2003–2005; and the Sleep Heart Health Study (SHHS), which enrolled 6,441 participants aged 40+ between 1995–1998 to investigate cardiovascular outcomes of sleep-disordered breathing. Additional datasets are summarized in \Cref{tab:psg_datasets}. 

However, existing collections have key limitations, including small cohorts, inconsistent modality availability, limited benchmarking tasks, and no standardized train/validation/test splits. \dataset{} addresses these gaps with a large-scale, structured framework for evaluating sleep foundation models.

%
%

\section{Results}
\label{section_experiments}

\subsection{Dataset and Preprocessing} 
{\dataset} comprises 17,467 overnight polysomnography (PSG) recordings collected from a large sleep center. Use and sharing of these PSGs were approved under IRB. Following standard clinical practice guidelines \cite{berry2012aasm}, we selected a consistent set of 16 channels across four key physiological modalities: BAS, RESP, EKG, and EMG. The BAS modality includes central and occipital brain activity and eye movements. RESP includes thoraccoabdominal effort, oxygen saturation, and airflow. Cardiac activity is captured with a bipolar EKG lead, while EMG measures muscle tone and leg movements.
To ensure signal quality and computational efficiency, all channels were resampled to 128 Hz using anti-aliasing filters, with zero-phase butterworth low-pass filters applied at the nyquist frequency. This sampling rate preserves essential physiological frequencies across all modalities. The dataset was randomly partitioned into training ($n=12,952$), validation ($n=1,500$), and test ($n=3,015$) sets, with subject-level separation maintained between splits. Additional details about demographics and modalities are provided in \Cref{tab:demographics} and Supplementary \Cref{tab:modalities}.

\begin{table*}[t!]
\centering
\caption{Demographics across different splits of \dataset. Additional details on sleep staging, apnea diagnosis, and disease prediction labels are available in \Cref{tab:demographics_outcomes}.}
\label{tab:demographics}
\renewcommand{\arraystretch}{1.2}
\resizebox{\textwidth}{!}{%
\begin{tabular}{llrrr}
\toprule
& & \textbf{Pre-train/Train} & \textbf{Test} & \textbf{Validation} \\
& & \textit{n} = 12,952, \textit{h} = 121,365 & \textit{n} = 3,015, \textit{h} = 28,306 & \textit{n} = 1,500, \textit{h} = 13,979 \\
\midrule
{\textbf{Age (years)}} & Mean ± SD & 43.27 ± 19.87 & 43.98 ± 19.58 & 43.02 ± 19.92 \\
\midrule
{\textbf{BMI (kg/m²)}} & Mean ± SD & 27.93 ± 7.56 & 27.73 ± 6.83 & 27.52 ± 6.92 \\
\midrule
{\textbf{Gender (n)}} 
& Male & 7,780 & 1,868 & 905 \\
& Female & 5,167 & 1,146 & 595 \\
& Unknown & 5 & 1 & 0 \\
\midrule
{\textbf{Race (n)}} 
& White & 7,314 & 1,756 & 868 \\
& Unknown & 2,582 & 560 & 297 \\
& Asian & 1,484 & 316 & 171 \\
& Other & 1,117 & 264 & 114 \\
& Black & 306 & 88 & 30 \\
& Pacific Islander & 120 & 17 & 17 \\
& Native American & 29 & 14 & 3 \\
\midrule
{\textbf{Ethnicity (n)}} 
& Non-Hispanic & 9,378 & 2,243 & 1,091 \\
& Unknown & 2,586 & 559 & 292 \\
& Hispanic/Latino & 988 & 213 & 117 \\
\bottomrule
\end{tabular}%
}
\end{table*}

\begin{table*}[t]
\centering
\caption{Summary of public PSG datasets with $>$ 1,000 multi-modal recordings. The table details modalities, benchmark tasks (where mortality and disease prediction refer to forecasting future occurrences), and reproducibility details. Only \dataset{} is structured as a standardized benchmark. * indicates partial availability. \dataset{} has 17,467 PSG distributed over 12,794 subjects.}
\renewcommand{\arraystretch}{1.2}
\resizebox{%
      \ifdim\width>\textwidth
        \textwidth
      \else
        \width
      \fi
    }{!}{%
\scriptsize
\begin{tabular}{l r cccc ccccc cc}
\toprule
{\textbf{Dataset}} & 
{\textbf{Subjects}} & 
\multicolumn{4}{c}{\textbf{Modalities}} & 
\multicolumn{5}{c}{\textbf{Benchmark}} & 
\multicolumn{2}{c}{\textbf{Reproducibility}} \\
\cmidrule(lr){3-6} \cmidrule(lr){7-11} \cmidrule(lr){12-13}
& & BAS & RESP & EKG & EMG & Sleep Stage & Apnea & Age Estimation & Mortality & Diseases & Codebase & Model Weights \\
\midrule
CHAT \cite{marcus2013randomized} & 1,243 & 12 & 6 & 3 & 3 & \checkmark & \checkmark & \checkmark & - & - &  - & - \\
APPLES \cite{quan2011association} & 1,516 & 7 & 8 & 1 & 1 & \checkmark & \checkmark & \checkmark & - & - & - & -  \\
STAGES \cite{zhang2018national} & 1,881 & 4+ & 5+ & 1+ & 1+ & \checkmark & \checkmark & \checkmark & - & - & -  & - \\
MESA \cite{chen2015racial} & 2,237 & 5 & 5 & 1 & 2 & \checkmark & \checkmark & \checkmark & \checkmark & *  & - & - \\
MrOS \cite{blackwell2011associations} & 2,911 & 6 & 5 & 2 & 3 & \checkmark & \checkmark & \checkmark & \checkmark & *  & - & - \\
MNC \cite{stephansen2018neural} & 3,000 & 11 & 6 & 3 & 2 & \checkmark & - & - & - & - & \checkmark & \checkmark \\
NCHSDB \cite{lee2022large} & 3,673 & 9 & 5 & 1 & 2 & \checkmark & \checkmark & \checkmark & - & \checkmark & -  & - \\
SHHS \cite{quan1997sleep} & 5,804 & 4 & 4 & 1 & 0 & \checkmark & \checkmark & \checkmark & \checkmark & *  & - & - \\
HSPv2 \cite{westover2023human} & 19,492 & 6+ & 3+ & 1+ & 3+ & \checkmark & \checkmark & \checkmark & - & -  & - & - \\
\midrule
\textbf{\dataset}\cite{kjaer2025stanfordsleepbench} & \textbf{12,794} & 8 & 5 & 1 & 2 & \checkmark & \checkmark & \checkmark & \checkmark & \checkmark & \checkmark & \checkmark \\
\bottomrule
\end{tabular}
}
\label{tab:psg_datasets}
\end{table*}

\subsection{Experiments}
\label{subsection_experiments}
We evaluated our models on five downstream task groups---sleep staging, apnea diagnosis, age estimation, and disease and mortality prediction---comparing their performance after fine-tuning on the full training set, along with in a few-shot setting.
During pre-training, we allowed all models to converge before fine-tuning them on downstream tasks for assessment.

\begin{figure*}[b!]
    \centering
    \includegraphics[width=\linewidth]{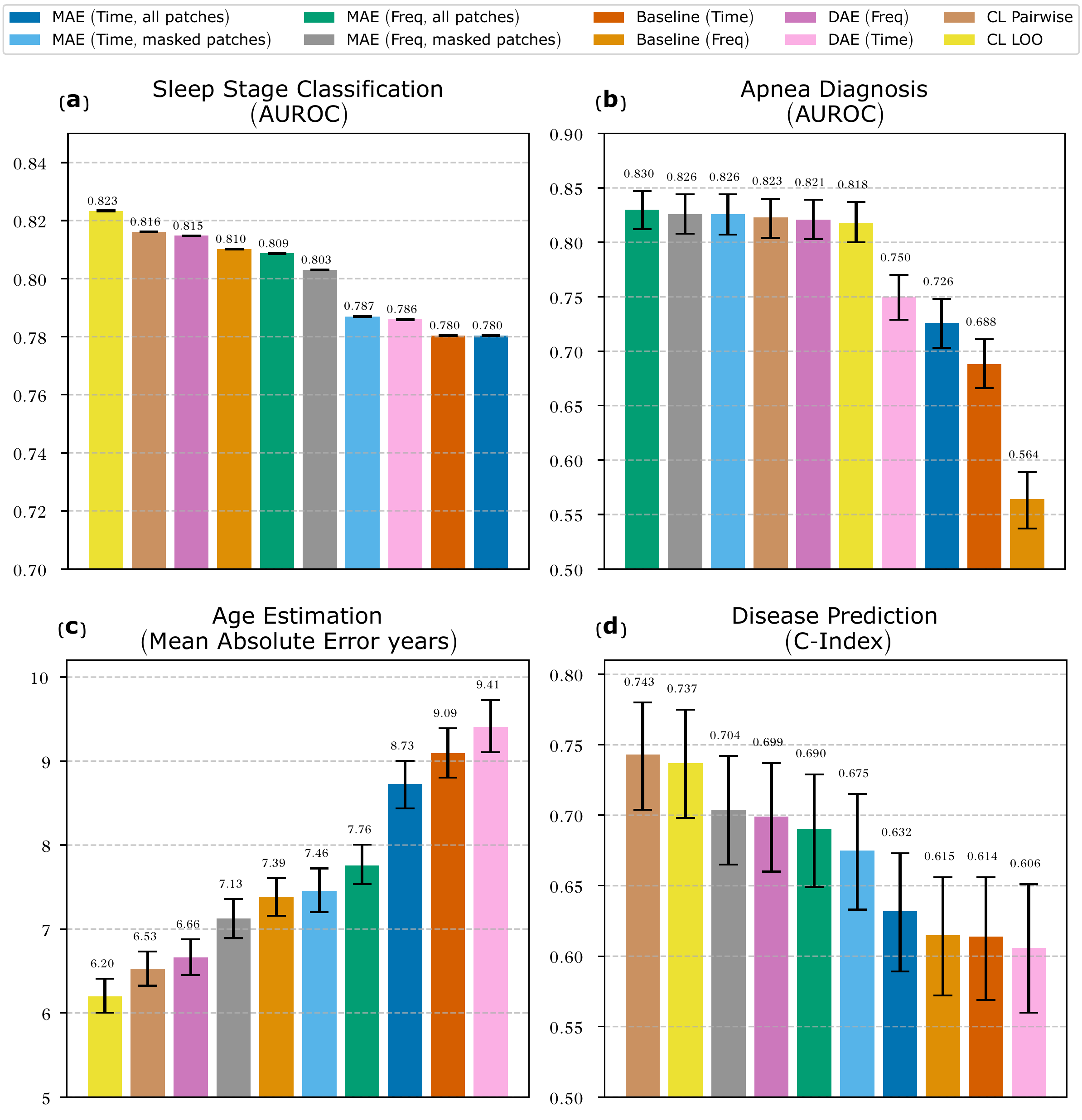}
    \caption{Comparison of self-supervised representation learning methods across sleep staging, apnea diagnosis, age estimation, and overall disease prediction, including mortality.}
    \label{fig:main results}
\end{figure*}

As our main results, we fine-tuned the models on the entire training set.
Our results indicate that CL, particularly the CL-LOO and CL-Pairwise approaches, consistently outperforms other methods, making it the most effective strategy for disease and mortality prediction (\Cref{fig:main results}). While reconstruction-based methods such as MAE and DAE achieve competitive performance in the frequency domain, their time-domain variants exhibit only marginal improvements over baselines, suggesting that the frequency domain objectives better capture sleep-related features.

To assess few-shot learning, we trained models on three replicates of 1 to 1024 datapoints and evaluated them on the test set. CL demonstrated superior efficiency in this setting, reaching 95\% of its best sleep staging performance with only 64 subjects, whereas baseline methods required significantly more data to achieve comparable accuracy.


\noindent\textbf{Main Results} \hspace{0.5em}
The best-performing sleep staging results across all pre-trained models are presented in \Cref{fig:main results} (a). Overall, contrastive learning methods achieve the highest performance, with CL-LOO (\(0.823\)) and CL-Pairwise (\(0.816\)). Following closely, the next best performances are obtained with DAE (Freq) (\(0.815\)) and MAE (Freq, all patches) (\(0.809\)). Surprisingly, Baseline (Freq) also performs strongly, achieving an AUROC of \(0.810\), comparable to more sophisticated pre-training methods. Additionally, MAE (Freq, masked patches) achieves a notable performance of \(0.803\). In contrast, time-domain methods show lower performance, with MAE (Time, masked patches) achieving \(0.787\), followed closely by DAE (Time) at \(0.786\), while MAE (Time, all patches) and Baseline (Time) yield slightly lower performances at \(0.780\). These results highlight the strengths of frequency-domain methods, which consistently outperform their time-domain counterparts, as well as the effectiveness of contrastive learning pre-training objectives in achieving robust performance. A detailed performance breakdown across all sleep stages is provided in Supplementary \Cref{tab:sleep_stage_results}.

\begin{figure*}[!t]
    \centering
    \includegraphics[width=\linewidth]{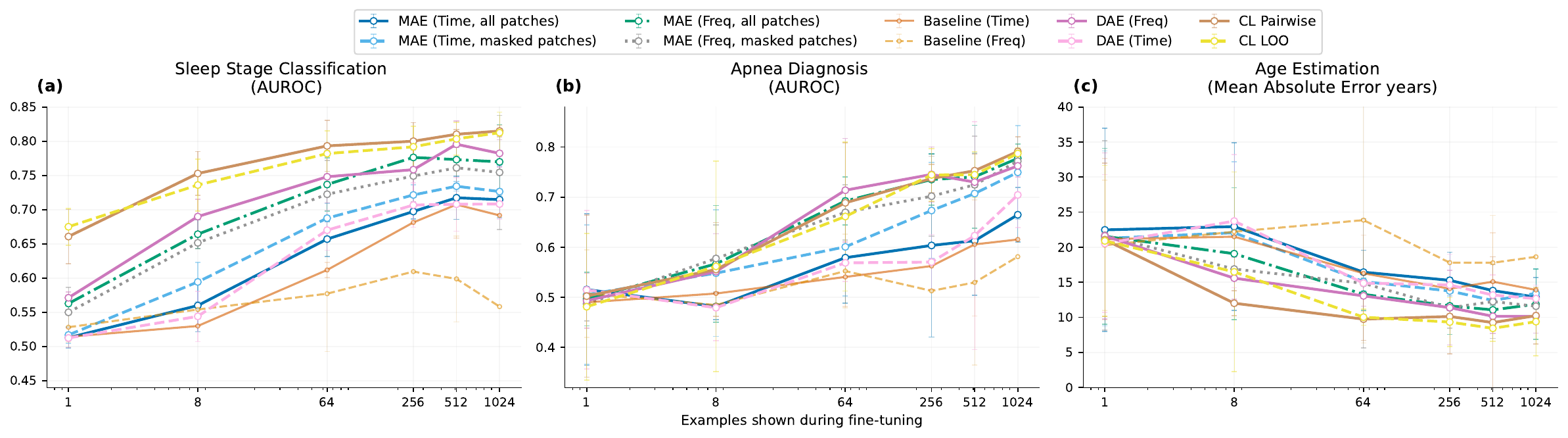}
    \caption{Few-shot performance of self-supervised representation learning methods on three replicates of [1, 8, 64, 256, 512, 1024] subjects across sleep staging, apnea diagnosis, and age estimation.}

    \label{fig:few shot results}
\end{figure*}

Sleep apnea diagnosis results are presented in \Cref{fig:main results} (b). MAE in the frequency domain achieves the highest performance, with full reconstruction reaching an AUROC of \(0.830\) and masked patches slightly lower at \(0.826\). CL methods perform comparably well, with CL-Pairwise achieving \(0.823\) and CL-LOO following closely at \(0.818\). In the time domain, MAE with masked patch reconstruction significantly outperforms full reconstruction, achieving an AUROC of \(0.826\) compared to \(0.726\). DAE in the frequency domain remains competitive with an AUROC of \(0.821\), although its time-domain counterpart drops to \(0.750\). In contrast, baselines perform substantially worse than SSRL methods, with the frequency baseline achieving \(0.564\) and the time-domain baseline \(0.688\).

Age estimation results reveal three distinct performance tiers as shown in \Cref{fig:main results} (c). The top tier comprises CL-LOO, CL-pairwise, and DAE (Freq), achieving near state-of-the-art performance \cite{brink2022age} (who achieved 5.8 ± 1.6 years) in wider range of ages. CL-LOO demonstrates the best accuracy with a mean absolute error of $6.20$ years, followed closely by CL Pairwise at $6.53$ years and DAE (Freq) at $6.67$ years. The second tier includes MAE (Freq, masked patches), Baseline (Freq), MAE (Time, masked patches), and MAE (Freq, all patches), which exhibit moderately lower accuracy. MAE (Freq, masked patches) achieves a mean error of $7.13$ years, while Baseline (Freq) and MAE (Time, masked patches) perform similarly at $7.39$ and $7.46$ years, respectively. MAE (Freq, all patches) records a slightly higher error of $7.76$ years. The last tier consists of MAE (Time, all patches), Baseline (Time), and DAE (Time), which show substantially reduced performance. MAE (Time, all patches) yields a mean error of $8.73$ years, while Baseline (Time) and DAE (Time) perform even worse, with errors of $9.10$ and $9.41$ years, respectively. Age estimation has the lowest error in young subjects across all models and the main difference in the models that perform well is on estimating the age of older subjects (Supplementary \Cref{fig:age stratified}).

Similarly, disease and mortality prediction results reveal three distinct performance tiers across pre-training methods. Contrastive learning approaches achieve the highest average performance across all twelwe diseases and mortality prediction, with CL-LOO and CL-pairwise both reaching C-indices of $0.74$. The second tier comprises frequency-domain methods, with MAE (Freq, masked patches) and DAE (Freq) both achieving C-indices of $0.70$, followed by MAE (Freq, all patches) at $0.69$ and MAE (Time, masked patches) at $0.68$. Full reconstruction in the time-domain methods form the lowest tier, with C-indices ranging from $0.64$ to $0.62$. This performance hierarchy remains consistent across individual disease categories and mortality prediction (Supplementary \Cref{tab:diagnosis_results}), suggesting that the advantages of contrastive learning and frequency-domain processing across clinical prediction tasks.

We compared the pre-training methods in a compute-controlled evaluation, where we pre-trained the model for 1, 4, and 16 epochs before fine-tuning and evaluating their preformance. Across training epochs, CL consistently outperformed other methods. However, all models converged on downstream task performance after as few as four pre-training epochs, as shown in Appendix Figure \ref{fig:learning_curves}.

\noindent\textbf{Few-Shot comparison} \hspace{0.5em}
For sleep staging, age estimation, and apnea diagnosis, we evalauted each SSRL method's ability to perform few-shot learning using three replicates of 1, 8, 64, 256, 512, and 1024 subjects. The results are presented in \Cref{fig:few shot results}. In sleep staging task, CL-pariwise demonstrates strong few-shot learning capabilities, reaching $\sim$ 95\% of its best sleep staging performance as reported in \Cref{fig:main results} (a), with only 64 subjects, while CL-LOO closely follows at 94\% at 64 subjects. DAE(Freq), MAE(Freq, all patches), MAE(Freq, masked patches) also show rapid few-shot learning but reach only 90\% of its top-performance at 64 subjects and fail to reach 95\% even at 1024 subjects. MAE(Time, all/masked patches) and DAE(Time) perform marginally better than the time-domain baseline. Neither baseline is able to learn sleep staging to a sufficient level within 1024 subjects.

For apnea detection, the six top performing SSRL methods on the full dataset show similar few-shot trends, though MAE (Time, Masked) learns more slowly. As shown in \Cref{fig:few shot results} (b), achieving 94\% performance requires 1024 examples, with lower performance for smaller sample sizes. Baseline are the slowest learners, while MAE (Time, all patches) and DAE (Time) perform slightly better than baseline. Similarly, for age estimation, CL-LOO and CL-pairwise demonstrates the fastest learning, while frequency-domain variations of DAE and MAE with masked reconstructions outperform those reconstructing all patches. MAE and DAE in the time domain perform similarly to baselines.


\noindent\textbf{External validation on SHHS} \hspace{0.5em}
To assess generalizability, we externally validate the SSRL methods on the SHHS dataset, yielding similar results to internal tests. As shown in \Cref{tab:shhs_results}, the best performing models are contrastive learning with masked autoencoding, optimized on masked segments in both the time and frequency domains, performing well on both sleep staging and outcome prediction.

\begin{table*}[t]
\centering
\caption{C-index for diagnosis and death prediction, and AUROC for sleep staging. Column headers: PW = Pairwise, LOO = Leave-One-Out, F = Freq, T = Time, M = Masked, A = All.}
\small
\resizebox{\textwidth}{!}{%
\begin{tabular}{lcccccccc}
\toprule
Model & CL PW & CL LOO & MAE (F, M) & DAE (F) & MAE (F, A) & MAE (T, M) & MAE (T, A) & DAE (T) \\
\midrule
Sleep staging  & 0.866 & 0.870 & 0.853 & 0.871 & 0.863 & 0.833 & 0.833 & 0.836 \\
\midrule
Average outcome  & 0.756 & 0.736 & 0.734 & 0.718 & 0.691 & 0.741 & 0.572 & 0.631 \\
Angina & 0.704 & 0.657 & 0.727 & 0.732 & 0.611 & 0.705 & 0.536 & 0.673 \\
CHD death & 0.803 & 0.820 & 0.771 & 0.749 & 0.788 & 0.769 & 0.613 & 0.654 \\
CHF & 0.778 & 0.761 & 0.763 & 0.735 & 0.717 & 0.774 & 0.578 & 0.657 \\
CVD death & 0.798 & 0.797 & 0.755 & 0.725 & 0.771 & 0.776 & 0.597 & 0.633 \\
MI & 0.698 & 0.660 & 0.674 & 0.672 & 0.572 & 0.685 & 0.528 & 0.575 \\
Stroke & 0.752 & 0.721 & 0.715 & 0.696 & 0.687 & 0.735 & 0.580 & 0.591 \\
\bottomrule
\end{tabular}%
}
\label{tab:shhs_results}
\end{table*}

%
%

\section{Discussion}
\label{section_discussion}

We introduced {\dataset}, a comprehensive PSG dataset and benchmark that aims to accelerate the development of sleep foundation models. With 17,467 PSG recordings spanning 163,650 hours and 10 years (2004-2014), {\dataset} represents one of the largest standardized sleep datasets to date. Its rigorous benchmarking framework, supporting 13 clinical prediction tasks alongside canonical sleep assessments, addresses a critical gap in the field where the lack of standardized evaluation has hindered systematic comparison of different modeling approaches. 

We comprehensively evaluated SSRL methods on {\dataset} and found several key insights. For canonical sleep tasks (sleep staging, apnea diagnosis), multiple approaches achieve comparable performance, with CL, frequency-domain MAE, and DAE all outperforming baseline embeddings. However, for complex tasks such as age estimation and disease prediction, which require integration of multimodal information across a PSG across an entire night, CL methods demonstrate superiority (6.2 years mean absolute error and a C-index of 0.743 for disease prediction), suggesting they better capture the complex multimodal relationships necessary for these predictions.

External validation on SHHS demonstrates that our findings generalize beyond our development dataset. The relative performance of SSRL approaches remained consistent despite differences in cohort demographics, technical protocols, and data collection period. This consistency among data sets with different characteristics suggests that observed performance differences reflect the properties of the learning methods rather than dataset-specific factors, supporting the potential applicability of these methods in diverse clinical settings.

A central barrier to progress in sleep AI is the lack of standardized evaluation, e.g., in sleep staging literature \cite{stephansen2018neural,perslev2021u,vallat2021open}, which makes cross-study comparisons difficult. Researchers often define bespoke splits, select different patient subsets, and report metrics on non-comparable cohorts. In sleep staging, several studies have shown that generalization across datasets is dependent on the expert's interpretation of the scoring guidelines in the training set \cite{Rosenberg2014TheEvents, liu2021large, olesen2021automatic}. {\dataset} addresses this gap by providing fixed data partitions and common evaluation protocols to enable direct comparisons. Beyond canonical tasks, the benchmark includes prospective follow-up for 12 diseases and all-cause mortality (13 outcomes), enabling investigation of how sleep relates to future disease risk. We found that the inclusion of complex tasks was imperative in evaluating and comparing SSRL methods, which underscores the need for {\dataset}. Models that surface early markers of cardiovascular, metabolic, or neurodegenerative disease could support risk stratification and preventive care. Notably, beyond traditional preventative measures of such diseases, sleep is modifiable through behavioral and pharmaceutical interventions \cite{wunderlin2020role}.


{\dataset} is complementary to other large data repositories such as the NSRR \cite{nsrr_matrix} and HSP \cite{westover2023human} and provides standardized splits and tooling for rapid, controlled ablations. Models developed on {\dataset} should be evaluated for generalizability on independent cohorts, like SHHS\cite{quan1997sleep}, to account for demographic, instrumentation, and scoring differences, which has been shown to be a prevailing issue in sleep staging \cite{olesen2021automatic}. Disease-specific datasets further complement population-based cohorts by enabling targeted evaluation of phenotype-specific outcomes (e.g., cardiovascular, metabolic, and neurodegenerative conditions).



The superior performance of CL relative to MAE, DAE, and baseline models likely stems from the combination of its temporal attention pooling and discriminative objective. The attention pooling aggregates information across 5-minute spans before the contrastive loss is applied, enabling embeddings to capture long-range and cross-modal physiological dependencies rather than focusing on local signal fidelity. In contrast, reconstruction-based approaches such as MAE and DAE primarily optimize for accurate short-term signal recovery, which preserves local waveform details but may fail to encode long-range physiological dynamics necessary for clinical prediction. Nonetheless, CL constrains its representation learning to information that is shared across modalities or temporal views, potentially neglecting modality-specific patterns that could hold diagnostic value. For instance, slow oscillation spindle-coupling strength, which is a marker of cognition and aging, is unlikely to be temporally coupled with the remaining modalities \cite{hahn2020slow}. Therefore, hybrid strategies combining contrastive and generative objectives or incorporating modality-specific auxiliary losses could further enrich learned representations and improve downstream generalization.

Our systematic comparison of pre-training objectives addresses only one dimension of model development. Architecture, optimization, and fine-tuning strategies are additional tunable dimensions of model performance. {\dataset} enables controlled ablations along these axes under consistent evaluation, facilitating clear attribution of gains to specific design choices. The modular codebase further supports such studies by allowing targeted substitutions of components. Accordingly, we see three immediate avenues for future work with {\dataset}: (i) expanding and refining SSRL objectives; (ii) exploring architectural variations tailored to multimodal PSG; and (iii) developing fine-tuning strategies robust to the sparsity of rare diagnoses.

We anticipate that {\dataset} will accelerate clinically meaningful PSG modeling by enabling methods explicitly optimized for the multimodal biosignals present in overnight studies, thus improving both performance of canonical sleep tasks and disease risk.





Despite these contributions, limitations of our work must be acknowledged. First, we did not exhaustively search the parameter space due to computational constraints, leaving room for further improvements. Second, while we focused on widely used pretraining methods for biosignal data, there may be other promising approaches yet to be explored, which we aim to address in future work. Lastly, while our work primarily focuses on the sleep domain, we believe these methods will generalize to other fields with similar signal types, potentially amplifying their impact.


%


%

\section{Method}
\label{section_method}


\noindent\textbf{Downstream Tasks} \hspace{0.5em}
To comprehensively evaluate our learned representations, we examined five downstream task categories of varying complexity (label distributions in Supplementary \Cref{tab:demographics_outcomes}).

For clinical sleep analysis, we evaluated two primary tasks. \textbf{Sleep staging} classifies 5-second patches into one of five sleep stages (Wake, N1, N2, N3, REM), following trends in automatic sleep staging methods \cite{Stephansen2018NeuralNarcolepsyb}, offering finer granularity than AASM guidelines \cite{berry2012aasm}. \textbf{Sleep apnea diagnosis} is based on classifying subjects according to their apnea-hypopnea index (AHI), with AHI $\geq$ 15 indicating sleep apnea. These tasks assess the model's ability to capture both spectral patterns in brain activity and respiratory events with their physiological responses.

Beyond these, we introduce more complex downstream tasks requiring multimodal signal integration. \textbf{Age estimation} is formulated as a regression problem, where performance serves as a proxy for capturing multimodal age-related physiological changes in sleep architecture, including alterations in brain, breathing, cardiac, and movement patterns. Finally, \textbf{mortality} and \textbf{disease prediction} as a time-to-event modeling task spanning cardiovascular, respiratory, metabolic, and neurological domains. This task evaluates whether the learned embeddings encode early clinically relevant markers of disease. Mortality and disease prediction are the most challenging tasks in \dataset{}, as the relationships between sleep physiology and disease risk remain poorly understood.

\noindent\textbf{Common Architecture} \hspace{0.5em}
All models share a common base architecture to generate embeddings, which are used to optimize their respective pretraining objectives. The architecture operates on 300-second segments of 16-channel PSG data, each channel is further divided into 5-second patches and passed through the corresponding modality-specific encoder (see \Cref{fig:overview}). The patch duration was chosen based on two key considerations: (1) ensuring computational feasibility when processing up to 8 hours of high-frequency data for downstream tasks and (2) empirical evidence indicating that smaller window sizes enhance temporal resolution for disease diagnosis \cite{Stephansen2018NeuralNarcolepsyb}.

Each modality is encoded by a modality-specific CNN-based patch encoder projecting the 8 BAS, 5 RESP, 1 EKG, and 2 EMG channels into an embedding in $\mathbb{R}^{128}$ with an average reduction ratio of 20:1. Positional encoding \cite{vaswani2017attention} is added before all modalities are independently processed by a transformer that captures temporal dependencies across patches. This architecture ensures all pre-training tasks can be compared with the common base architecture. Finally, embeddings from all modalities are concatenated into $\mathbb{R}^{512}$. An overview can be seen in \Cref{fig:overview}.

\noindent\textbf{Pre-training Schemes} \hspace{0.5em}
We evaluated variants of MAE, DAE, and CL, and two baseline methods.


\noindent\textbf{\textit{Baseline Embeddings}} \hspace{0.5em}
\textbf{Baseline (Time)} selects one representative channel from each modality group: $C3-M2$ (BAS), $NASAL$ (RESP), $EKG_L-EKG_R$ (EKG), and $LEG$ (EMG). These signals are down-sampled from 128 Hz to 25.6, providing a 5-fold reduction in temporal resolution to form a representation in $\mathbb{R}^{512}$ per 5 seconds to match the size of the other methods. \textbf{Baseline (Freq)} exploits the spectral characteristics of physiological signals through a three-step process. First, the discrete Fourier transform is applied to each channel to obtain its real frequency representation, excluding the Nyquist frequency. Second, within each modality group, frequency-domain representations are averaged across channels to generate modality-specific spectral signatures. Finally, 64 frequency amplitude and phase values are uniformly sampled from each modality's average spectrum. The features of all modalities are concatenated, forming a spectral representation in $\mathbb{R}^{512}$ per 5 seconds.

\noindent\textbf{\textit{Masked Auto-Encoder (MAE)}} \hspace{0.5em}
Our MAE implementation builds upon the common architecture by incorporating a lightweight decoder, similar to \cite{goswami2024moment}, consisting of a single fully connected layer that reconstructs all input signals from the learned embeddings. We explored four variations of the MAE approach: (1) \textbf{MAE (Time, all patches)}, which reconstructs both masked and unmasked patches in the time domain; (2) \textbf{MAE (Time, masked patches)}, which reconstructs only the masked patches in the time domain; (3) \textbf{MAE (Freq, all patches)}, which reconstructs both masked and unmasked patches in the frequency domain; and (4) \textbf{MAE (Freq, masked patches)}, which reconstructs only the masked patches in the frequency domain. The masking strategy randomly occludes 34\% of patches across both temporal and modality dimensions, maintaining this ratio consistently across all variants. Masking was applied independently to each channel. Across all MAE variants, the time-domain signals were used as input.

The reconstruction objectives are formulated using two complementary loss functions. In the time-domain, the mean squared error was used as the loss function \(\mathcal{L}_{\text{time}}\). For frequency-domain reconstruction, let \( X = \mathcal{F}(x) \) denote the discrete Fourier transform (DFT) of the target signal. We decompose the loss into amplitude and phase components to handle these distinct aspects of the frequency representation. To address differences in dynamic ranges of various physiological signals, we transform the amplitude spectrum using logarithmic scaling with an additive shift.

\begin{equation}
    A_X = \log_{10} (|X| + \epsilon) + C,
\end{equation}

where \( \epsilon = 10^{-6} \) is a small constant ensuring numerical stability, and \( C = -\log_{10}(\epsilon) \) is an additive shift to adjust the amplitude range to be non-negative. This transformation ensures that frequencies with zero amplitude do not result in negative infinity values, making the loss computation numerically stable. The amplitude loss \( \mathcal{L}_{\text{amp}}\) was computed as the mean squared error with the target \(A_X\). To account for phase periodicity, the phase loss was computed with shifts of \( 0 \), \( \pm 2\pi \):

\begin{equation}
    \mathcal{L}_{\text{phase}} =  \frac{1}{|\mathcal{K}|} \sum_{k \in \mathcal{K}} \min_{\delta \in \{0, -2\pi, 2\pi\}} (\angle X_k - (\angle \hat{X}_k + \delta))^2,
\end{equation}

where \( \mathcal{K} \) includes all frequencies except the Nyquist frequency. This ensures that small angular differences do not result in disproportionately large errors due to phase wrapping. The total frequency-domain loss is given by:

\begin{equation}
    \mathcal{L}_{\text{freq}} = \mathcal{L}_{\text{amp}} + \mathcal{L}_{\text{phase}}.
\end{equation}

\noindent\textbf{\textit{Denoising Auto-Encoder (DAE)}} \hspace{0.5em}
The denoising approach extends our architecture by training it to reconstruct clean signals from artificially corrupted inputs. Using the same decoder architecture and loss functions (\(\mathcal{L}_{\text{time}}\) and \(\mathcal{L}_{\text{freq}}\)) as the MAE, we create two variants: \textbf{DAE (Freq)} that denoises in the frequency-domain, and \textbf{DAE (Time)} that denoises in the time-domain. Before signals are input to the encoder, we corrupt the signals using a structured noise injection process. Specifically, we introduce additive white Gaussian noise to the input signals. The noise level for each channel is sampled from a uniform distribution within the range [0.01, 0.3] of the maximum signal amplitude. Noise is injected to each channel separately, with each channel having a 50\% probability of having noise added, such that approximately half of the available channels undergo corruption. For a given channel, the noise is generated and added per 300-second segment.

\noindent\textbf{\textit{Contrastive Learning (CL)}} \hspace{0.5em}
We explored two CL frameworks for learning joint representations across modalities: standard pairwise CL and leave-one-out CL (CL-LOO) \cite{Thapa2024SleepFM:Signals}. CL optimizes embeddings by bringing positive pairs closer in the latent space while pushing negative pairs apart. Each 5-second patch yields a vector in $\mathbb{R}^{512}$ by concatenating modality embeddings. To compute contrastive loss, we average 60 such segments (300 seconds) into an embedding in $\mathbb{R}^{128}$ per modality. These embeddings are used during pretraining, while the original 5-second representations are retained for downstream tasks. Positive pairs are formed from time-aligned mean-pooled embeddings across modalities, with non-matching batch samples as negatives. In \textbf{CL-pairwise}, contrastive loss is applied between all pairs of modalities:
\[
\mathcal{L}^{\text{pair}}_{i, j, k} = -\log \frac{\exp\left(\text{sim}(x_{k}^{i}, x_{k}^{j}) / \tau \right)}{\sum_{m=1}^{N} \exp\left(\text{sim}(x_{k}^{i}, x_{m}^{j}) / \tau \right)},
\]
where \( x_k^i \) and \( x_k^j \) are embeddings from modalities \( i \) and \( j \), \( N \) is the batch size, \( \tau \) is temperature, and \( \text{sim}(\cdot) \) is cosine similarity. The final loss averages over all modality pairs. In \textbf{CL-LOO}, we construct a leave-one-out representation \( \bar{x}_k^{-i} \) by averaging embeddings from all other modalities, excluding modality \( i \), and apply contrastive loss:
\[
\mathcal{L}^{\text{LOO}}_{i, k} = -\log \frac{\exp\left(\text{sim}(x_{k}^{i}, \bar{x}_k^{-i}) / \tau \right)}{\sum_{m=1}^{N} \exp\left(\text{sim}(x_{k}^{i}, \bar{x}_m^{-i}) / \tau \right)}.
\]

\noindent\textbf{Fine-tuning} \hspace{0.5em}
We fine-tuned all pre-trained models by freezing the common architecture and training a task-specific model. The fine-tuning model includes a two-layer transformer with four heads and a two-layer bidirectional LSTM. For 5-second patch-level tasks like sleep staging, a fully connected layer generates predictions. For full PSG recordings, including apnea diagnosis, age estimation, and disease prediction, an attention-pooling mechanism aggregates embeddings over eight hours using a single transformer layer with eight heads, followed by a classification layer for the final output.

We used cross-entropy loss for classification tasks including sleep staging and apnea diagnosis. For age estimation, we applied min-max normalization (scaling age so that 0 corresponds to 0 and 100 corresponds to 1) and used a softplus activation to prevent negative predictions, optimizing with mean squared error (MSE) loss. For death and disease time-to-event estimation, we used the Cox proportional hazards (CoxPH) loss function, defined as:

\[
\mathcal{L}_{\text{CoxPH}} = - \frac{1}{N_e} \sum_{i=1}^n \delta_i \left( h_i - \log \sum_{j \in R(t_i)} \exp(h_j) \right),
\]

where \( N_e = \sum_{i=1}^n \delta_i \) is the number of events, \(h_i\) is the predicted hazard for the \(i\)-th patient, \(\delta_i\) is the event indicator (1 if the event occurred, 0 otherwise), and \(R(t_i)\) represents the set of patients at risk at time \(t_i\). The CoxPH loss was computed independently for each of the 12 diseases, and the total disease loss was obtained by summing the individual losses, where \(L\) is the number of diseases.:

\[
\mathcal{L}_{\text{disease}} = \sum_{k=1}^{L} \mathcal{L}_{\text{CoxPH}}^{(k)},
\]

Additionally, the CoxPH loss for death prediction was computed separately. The final total loss, combining both disease and death prediction losses, is given by:

\[
\mathcal{L}_{\text{total}} = \mathcal{L}_{\text{disease}} + \mathcal{L}_{\text{death}}.
\]

\noindent\textbf{Implementation Details} \hspace{0.5em}
All methods share a common architecture processing 5-second patches (640 samples) through a temporal transformer (8 heads, 6 layers) with modality embeddings in $\mathbb{R}^{128}$. Models were pre-trained using Adam optimizer (learning rate 0.001, batch size 256) with early stopping on 100 validation PSGs. Training required 15-35 hours on 2 A100 GPUs. MAE randomly masked 34\% of patches, while DAE corrupted 50\% of input signals with noise.


\section{Data Availability}

Data is available through the \href{https://doi.org/10.60508/0ta5-v163}{Brain Data Science Platform}.

\section{Code Availability}

Code including benchmarking setup is shared on \href{https://github.com/RuudeResearch/SleepBench}{GitHub}. 

\bibliographystyle{plainnat}
\bibliography{sn-bibliography}

\section{Acknowledgments}

\section{Author Contributions Statement}

\section{Competing Interests Statement}


\appendix


\renewcommand{\figurename}{Supplementary Figure}
\renewcommand{\tablename}{Supplementary Table}

\clearpage
\section{Technical Appendices and Supplementary Material}
\label{section_appendix}

\subsection{Additional Data Explanation}
\label{subsection data explanation}

\begin{figure*}[h!]
    \centering
    \includegraphics[width=1\linewidth]{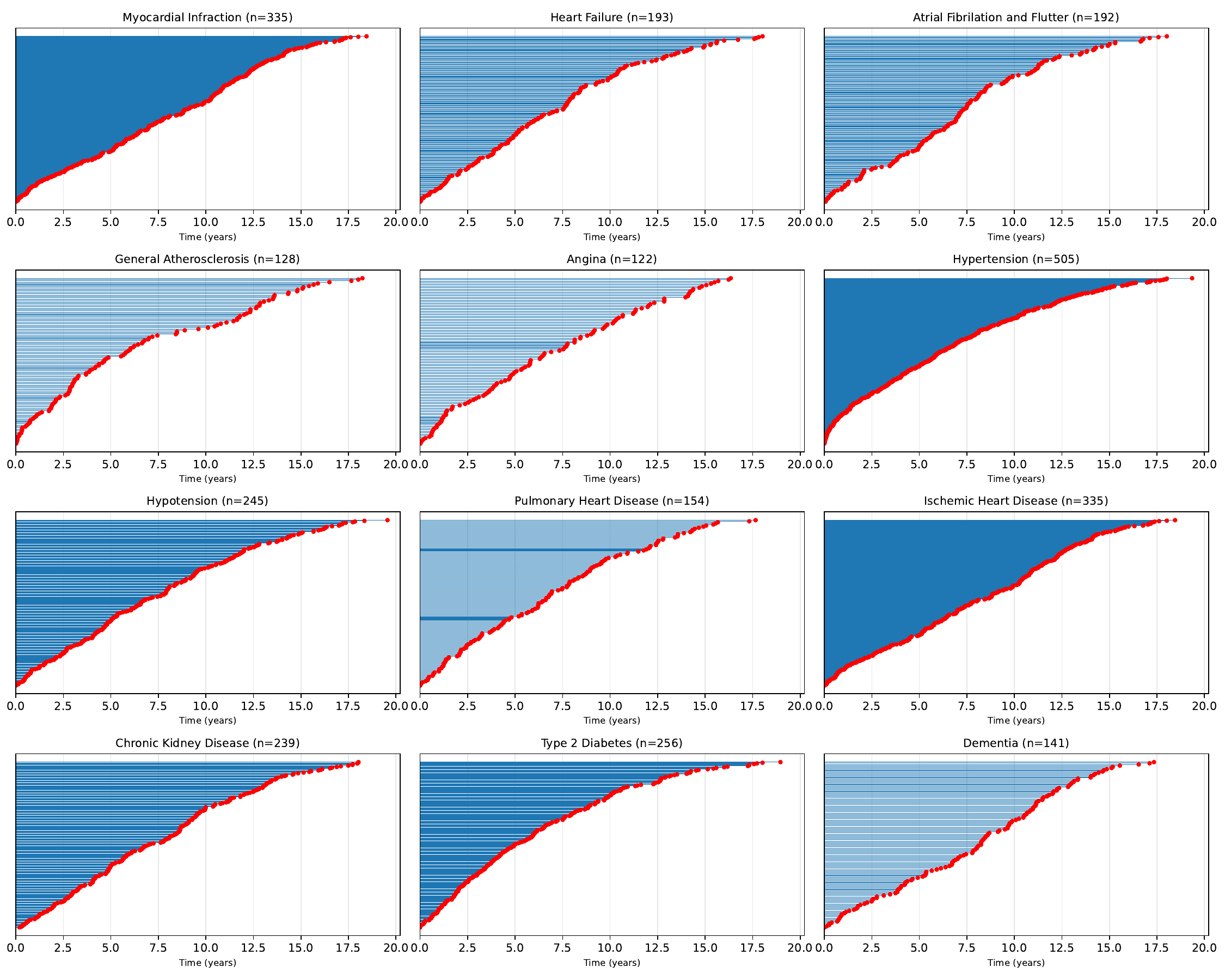}
    \caption{Swimmer plot of all 12 clinical conditions in {\dataset} on the test set, illustrating the timelines of positive cases from the date of PSG recording to the first occurrence of each condition.}
    \label{fig:overall_swimmer_plot_diseases}
\end{figure*}

\begin{figure*}[h!]
    \centering
    \includegraphics[width=1\linewidth]{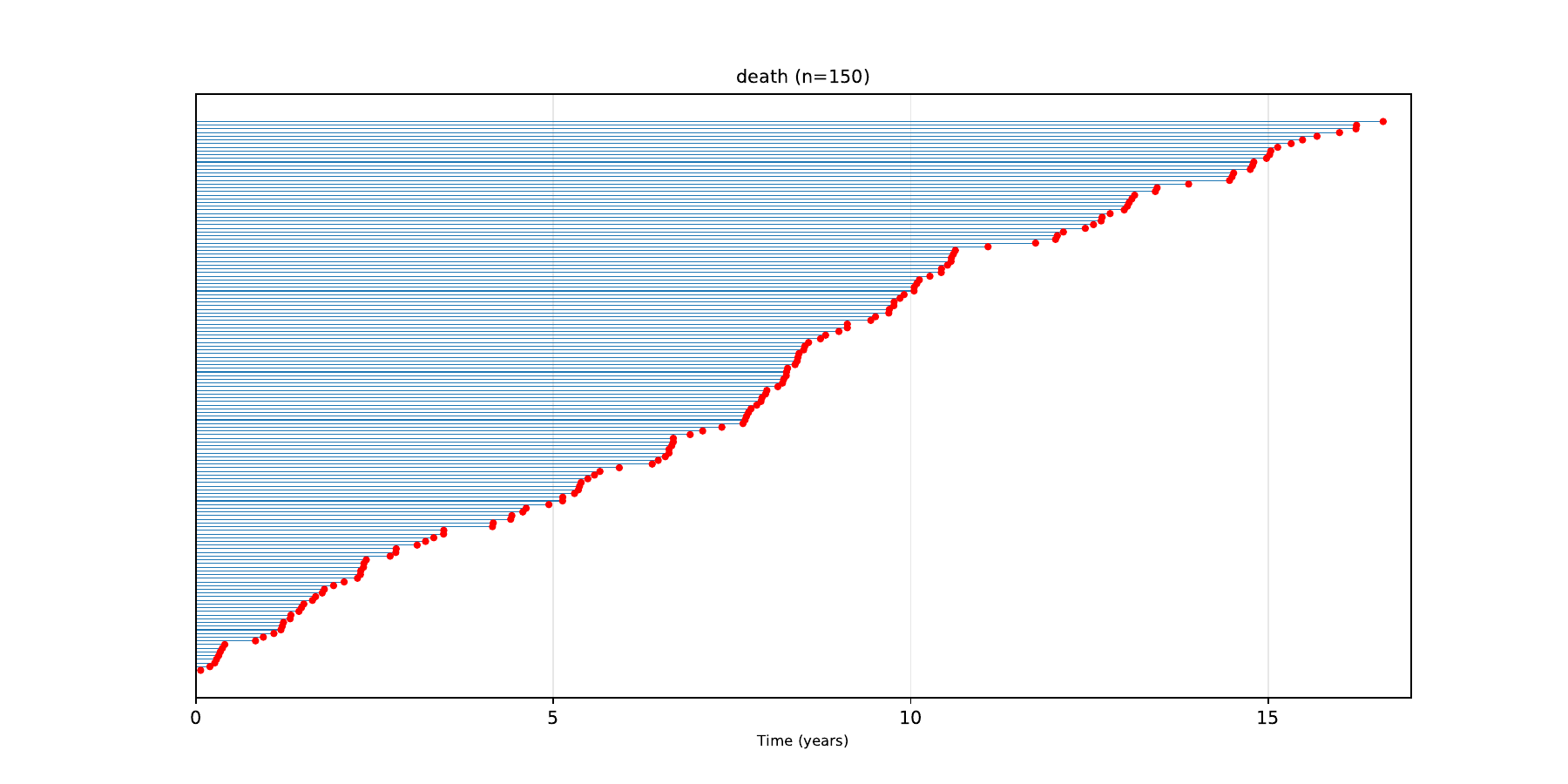}
    \caption{Swimmer plot of all-cause mortality in {\dataset} on the test set, showing the timeline from the date of PSG recording to the occurrence of death.}
    \label{fig:death_swimmer_plot}
\end{figure*}


\begin{table*}[!htbp]
    \centering
    \caption{Summary of modalities and their corresponding channels in {\dataset}.}
    \renewcommand{\arraystretch}{1.2}
    \begin{tabular}{|l|l|}
        \hline
        \textbf{Modality} & \textbf{Channels} \\ 
        \hline
        {BAS (Referential EEG)} 
        &$C3-M2$  (Left central brain activity) \\  
        &$C4-M1$  (Right central brain activity) \\  
        &$O1-M2$  (Left occipital brain activity) \\  
        &$O2-M1$  (Right occipital brain activity) \\  
        &$E1-M2$  (Left eye movement) \\  
        &$E2-M1$  (Right eye movement) \\  
        &$FP1-M2$ (Left prefrontal activity) \\  
        &$FP2-M1$ (Right prefrontal activity) \\          
        \hline
        {RESP Respiratory Monitoring} 
        & $CHEST$ (Thoracic Effort) \\
        & $SPO2$ (Oxygen Saturation) \\
        & $ABD$ (Abdominal Effort) \\
        & $NASAL$ (Nasal Airflow) \\
        & $ORAL$ (Oral Airflow) \\
        \hline
        {EKG Cardiac Activity} 
        & $EKG_L-EKG_R$ (Bipolar EKG Lead) \\
        \hline
        {EMG Muscle Activity} 
        & $CHIN$ (Submental EMG: avg$(\text{L}-\text{ctr}, \text{R}-\text{ctr})$)) \\
        & $LEG$ (Leg Movements: $RAT-LAT$) \\
        \hline
    \end{tabular}
    \label{tab:modalities}
\end{table*}

\begin{table*}[h!]
    \centering
    \caption{Task label distributions across different splits of \dataset. 
MI: Myocardial Infarction, HF: Heart Failure, AF: Atrial Fibrillation and Flutter, 
GA: General Atherosclerosis, AN: Angina, HT: Hypertension, HPT: Hypotension, 
PHD: Pulmonary Heart Disease, IHD: Ischemic Heart Disease, 
CKD: Chronic Kidney Disease, T2D: Type 2 Diabetes, DEM: Dementia.}
\scriptsize
\begin{tabular}{lccc}
\toprule
 & Pretrain/Train & Test & Validation \\
 & \textit{n} = 12,952 & \textit{n} = 3,015 & \textit{n} = 1,500 \\
  & \textit{h} = 121,365 & \textit{h} = 28,306 & \textit{h} = 13,979\\
\midrule
\textbf{Sleep Stages (h)} &&& \\
Wake & 1.82 ± 1.22 & 1.83 ± 1.20 & 1.83 ± 1.20 \\
N1 & 0.53 ± 0.47 & 0.53 ± 0.49 & 0.52 ± 0.47 \\
N2 & 4.07 ± 1.21 & 4.08 ± 1.19 & 4.10 ± 1.25 \\
N3 & 0.61 ± 0.70 & 0.57 ± 0.68 & 0.61 ± 0.67 \\
REM & 1.05 ± 0.57 & 1.05 ± 0.55 & 1.06 ± 0.58 \\
\midrule
\textbf{AHI} &&& \\
Mean ± SD & 21.59 ± 17.84 & 20.82 ± 17.04 & 21.34 ± 17.36 \\
\midrule
\textbf{Mortality (n)} &&& \\
Alive & 12227 & 2854 & 1417 \\
Deceased & 686 & 154 & 75 \\
\midrule
\textbf{Diseases (n)} &&& \\
MI & 1235 & 335 & 140 \\
HF & 696 & 193 & 73 \\
AF & 752 & 192 & 90 \\
GA & 336 & 128 & 48 \\
AN & 350 & 122 & 43 \\
HT & 1827 & 505 & 228 \\
HPT & 862 & 245 & 88 \\
PHD & 518 & 154 & 59 \\
IHD & 1235 & 335 & 140 \\
CKD & 875 & 239 & 90 \\
T2D & 985 & 256 & 109 \\
DEM & 575 & 141 & 72 \\
\bottomrule
\end{tabular}
\normalsize
\label{tab:demographics_outcomes}
\end{table*}

\subsection{Additional Results}
\label{subsection additional results}

\subsection{Stratified Sleep Staging Performance}

We show the stratified performance on sleep staging stratified on sleep stages in table \ref{tab:sleep_stage_results}.

\begin{table*}[htbp]
\centering
\caption{AUROC performance for sleep stage classification across models.}
\begin{tabular}{lccccc}
\toprule
Model & Wake & N1 & N2 & N3 & REM \\
\midrule

CL LOO & 86.7$\pm$0.0 & 70.8$\pm$0.1 & 84.6$\pm$0.0 & 77.0$\pm$0.0 & 92.5$\pm$0.0 \\
CL Pairwise & \textbf{87.6$\pm$0.0} & 68.8$\pm$0.0 & \textbf{85.6$\pm$0.0} & 72.7$\pm$0.0 & 93.3$\pm$0.0 \\
DAE(Fourier) & 86.2$\pm$0.0 & 69.4$\pm$0.0 & 84.8$\pm$0.0 & 74.4$\pm$0.0 & 92.6$\pm$0.0 \\
DAE (Time) & 81.6$\pm$0.0 & 64.3$\pm$0.0 & 81.7$\pm$0.0 & 75.5$\pm$0.0 & 89.9$\pm$0.0 \\
MAE(Fourier, all patches) & 82.9$\pm$0.0 & \textbf{71.3$\pm$0.0} & 84.8$\pm$0.0 & 71.9$\pm$0.0 & \textbf{93.4$\pm$0.0} \\
MAE(Fourier, masked patches) & 86.9$\pm$0.0 & 64.8$\pm$0.0 & 84.3$\pm$0.0 & 74.1$\pm$0.1 & 91.4$\pm$0.0 \\
MAE(Time, all patches) & 80.5$\pm$0.0 & 65.2$\pm$0.0 & 83.0$\pm$0.0 & 70.9$\pm$0.1 & 90.6$\pm$0.0 \\
MAE(Time, masked patches) & 82.6$\pm$0.0 & 67.4$\pm$0.0 & 83.8$\pm$0.0 & 69.1$\pm$0.0 & 90.5$\pm$0.0 \\
Baseline (Fourier) & 84.9$\pm$0.0 & 70.1$\pm$0.0 & 84.5$\pm$0.0 & 72.5$\pm$0.1 & 93.1$\pm$0.0 \\
Baseline (Time) & 80.5$\pm$0.0 & 62.6$\pm$0.0 & 81.3$\pm$0.0 & \textbf{77.4$\pm$0.0} & 88.4$\pm$0.0 \\

\bottomrule
\end{tabular}
\label{tab:sleep_stage_results}
\end{table*}

\subsubsection{Stratified Age Estimation}

\begin{figure*}[t]
    \centering
    \includegraphics[width=0.7\linewidth]{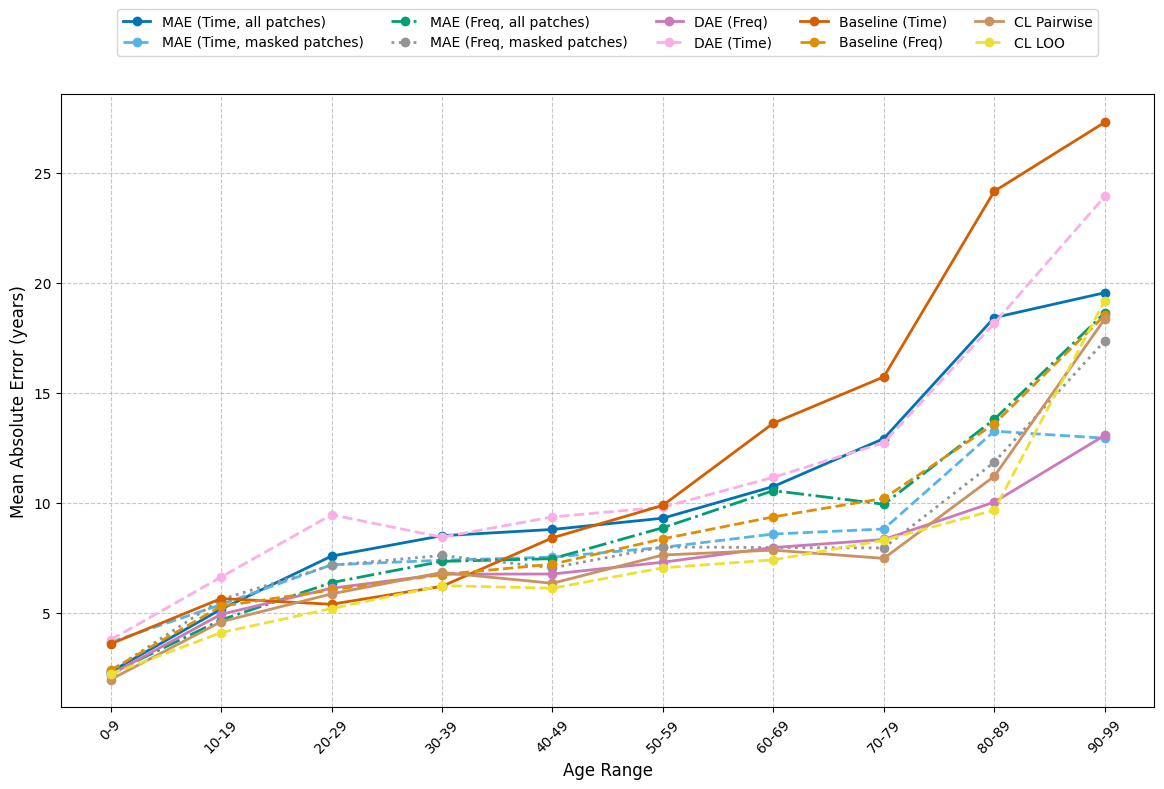}
    \caption{Age estimation error stratified on age.}
    \label{fig:age stratified}
\end{figure*}

We stratify the age estimation performance over target age in Figure \ref{fig:age stratified}.

In Supplementary \Cref{tab:diagnosis_results} we show the C-indexed for all diagnosis and mortality prediction separately. We visualize these results in Supplementary \Cref{fig:diagnosis}.

\begin{table*}[htbp]
\centering
\caption{C-index results for different models and modalities sorted left for the best overall diagnosis and death prediction performance. Column headers map to: CL PW (CL Pairwise), CL LOO (CL Leave-One-Out), MAE (F, M) (MAE Fourier, masked patches), DAE (F) (DAE Fourier), MAE (F, M) (MAE Fourier, all patches), MAE (T, M) (MAE Time, masked patches), MAE (T, A) (MAE Time, all patches), B (T) (Baseline Time), DAE (T) (DAE Time), and B (F) (Baseline Fourier). Diagnosis abbreviations: OVR: Overall, DTH: Death, D\textsubscript{OVR}: Diagnosis Overall, ANG: Angina, AFF: Atrial Fibrillation and Flutter, CKD: Chronic Kidney Disease, DEM: Dementia, GATH: General Atherosclerosis, HF: Heart Failure, HTN: Hypertension, HYPO: Hypotension, IHD: Ischemic Heart Disease, MI: Myocardial Infarction, PHD: Pulmonary Heart Disease, T2D: Type 2 Diabetes.}
\tiny
\resizebox{\textwidth}{!}{
\begin{tabular}{lcccccccccc}
\toprule
Model & CL PW & CL LOO & MAE (F, M) & DAE (F) & MAE (F, A) & MAE (T, M) & MAE (T, A) & B (F) & B (T) & DAE (T) \\
\midrule
OVR  & .74 (.70-.78) & .74 (.70-.78) & .70 (.67-.74) & .70 (.66-.74) & .69 (.65-.73) & .68 (.63-.72) & .63 (.59-0.67) & .62 (.57-.66) & .61 (.57-.66) & .61 (.56-.65) \\
DTH  & .84 (.80-.88) & .84 (.80-.88) & .79 (.74-.83) & .77 (.73-.81) & .75 (.69-.79) & .75 (.70-.79) & .68 (.63-0.73) & .61 (.56-.66) & .65 (.60-.70) & .65 (.60-.71) \\
D\textsubscript{OVR} & .73 (.70-.77) & .73 (.69-.77) & .70 (.66-.73) & .69 (.65-.73) & .69 (.65-.72) & .67 (.63-.71) & .63 (.58-0.67) & .61 (.57-.66) & .61 (.57-.65) & .60 (.56-.65) \\
ANG  & .70 (.65-.75) & .69 (.64-.74) & .67 (.63-.72) & .68 (.62-.73) & .68 (.62-.72) & .66 (.61-.72) & .59 (.54-.64) & .61 (.56-.66) & .63 (.57-.68) & .58 (.52-0.63) \\
AFF  & .73 (.69-.77) & .72 (.68-.76) & .68 (.65-.72) & .69 (.65-.72) & .67 (.63-.71) & .64 (.60-.69) & .62 (.58-.66) & .61 (.56-.65) & .56 (.51-.60) & .58 (.53-0.62) \\
CKD  & .75 (.72-.79) & .74 (.71-.78) & .71 (.67-.74) & .72 (.69-.76) & .69 (.65-.73) & .68 (.64-.72) & .64 (.60-.67) & .61 (.58-.65) & .63 (.59-.67) & .62 (.58-0.66) \\
DEM  & .81 (.77-.85) & .81 (.77-.85) & .76 (.72-.81) & .74 (.69-.78) & .76 (.72-.80) & .73 (.68-.77) & .68 (.63-.73) & .60 (.54-.65) & .62 (.56-.67) & .64 (.58-0.69) \\
GATH & .77 (.73-.81) & .77 (.74-.81) & .72 (.67-.76) & .73 (.68-.76) & .72 (.67-.76) & .69 (.65-.74) & .70 (.65-.74) & .65 (.60-.70) & .63 (.57-.67) & .68 (.63-0.72) \\
HF   & .79 (.75-.83) & .78 (.73-.81) & .72 (.69-.76) & .72 (.68-.76) & .71 (.68-.75) & .69 (.65-.73) & .64 (.60-.69) & .64 (.60-.69) & .63 (.59-.68) & .59 (.55-0.64) \\
HTN  & .71 (.69-.74) & .70 (.68-.72) & .69 (.66-.71) & .68 (.66-.71) & .70 (.67-.72) & .63 (.61-.66) & .59 (.56-.61) & .61 (.58-.63) & .58 (.55-.61) & .58 (.55-0.62) \\
HYPO & .66 (.62-.70) & .65 (.60-.69) & .64 (.60-.68) & .65 (.61-.69) & .64 (.60-.68) & .65 (.61-.69) & .60 (.55-.65) & .62 (.58-.66) & .62 (.58-.66) & .59 (.54-0.63) \\
IHD  & .73 (.70-.76) & .73 (.70-.77) & .70 (.67-.73) & .69 (.66-.72) & .68 (.64-.71) & .66 (.63-.70) & .63 (.60-.67) & .62 (.58-.65) & .61 (.57-.64) & .61 (.57-0.64) \\
MI   & .73 (.70-.76) & .73 (.70-.76) & .70 (.67-.73) & .69 (.66-.72) & .68 (.65-.71) & .66 (.63-.70) & .63 (.60-.67) & .62 (.59-.65) & .61 (.57-.64) & .61 (.57-0.64) \\
PHD  & .72 (.66-.77) & .72 (.67-.78) & .68 (.63-.73) & .67 (.62-.72) & .63 (.57-.68) & .67 (.62-.72) & .62 (.57-.67) & .61 (.56-.66) & .61 (.55-.66) & .58 (.52-0.64) \\
T2D  & .71 (.67-.74) & .70 (.66-.73) & .69 (.65-.72) & .67 (.63-.70) & .67 (.63-.71) & .64 (.60-.68) & .59 (.56-.63) & .58 (.54-.62) & .60 (.56-.64) & .57 (.53-0.61) \\
\bottomrule
\end{tabular}
}
\label{tab:diagnosis_results}
\end{table*}

\begin{figure*}[t!]
    \centering
    \includegraphics[width=\linewidth]{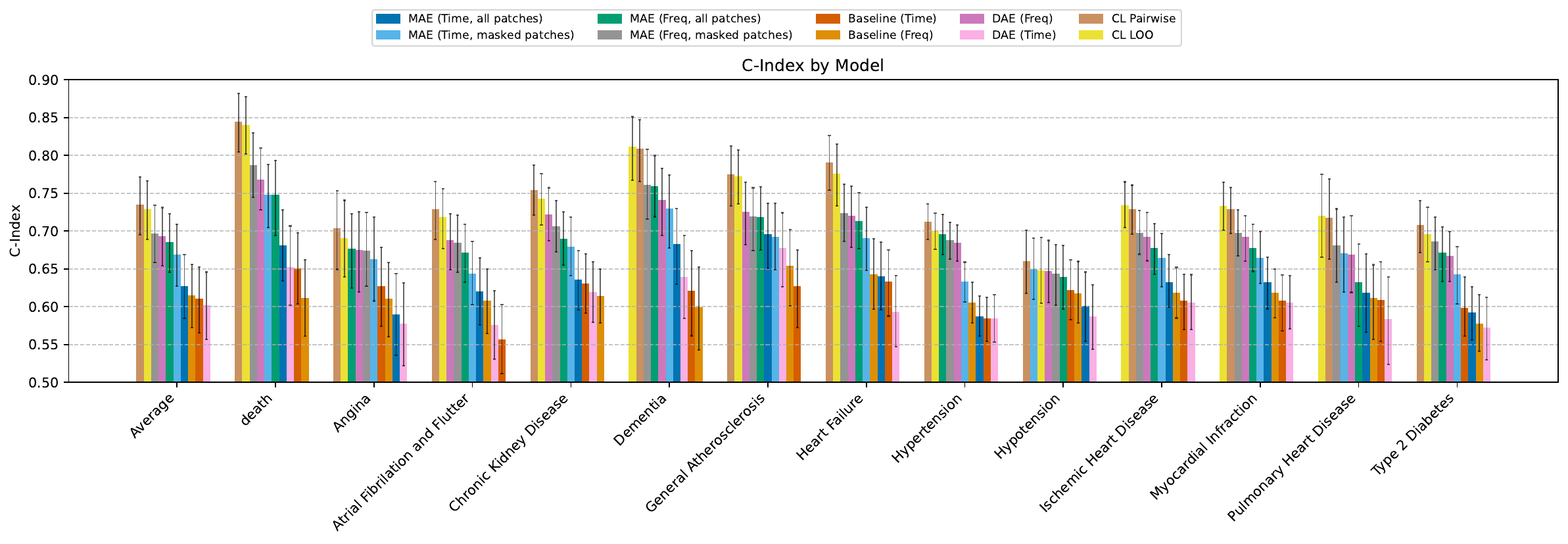}
    \caption{C-Index for death and diagnosis prediction, sorted from best to worst performing models.}
    \label{fig:diagnosis}
\end{figure*}

\subsubsection{Compute controlled performance}

\begin{figure*}[h!]
    \centering
    \includegraphics[width=\linewidth]{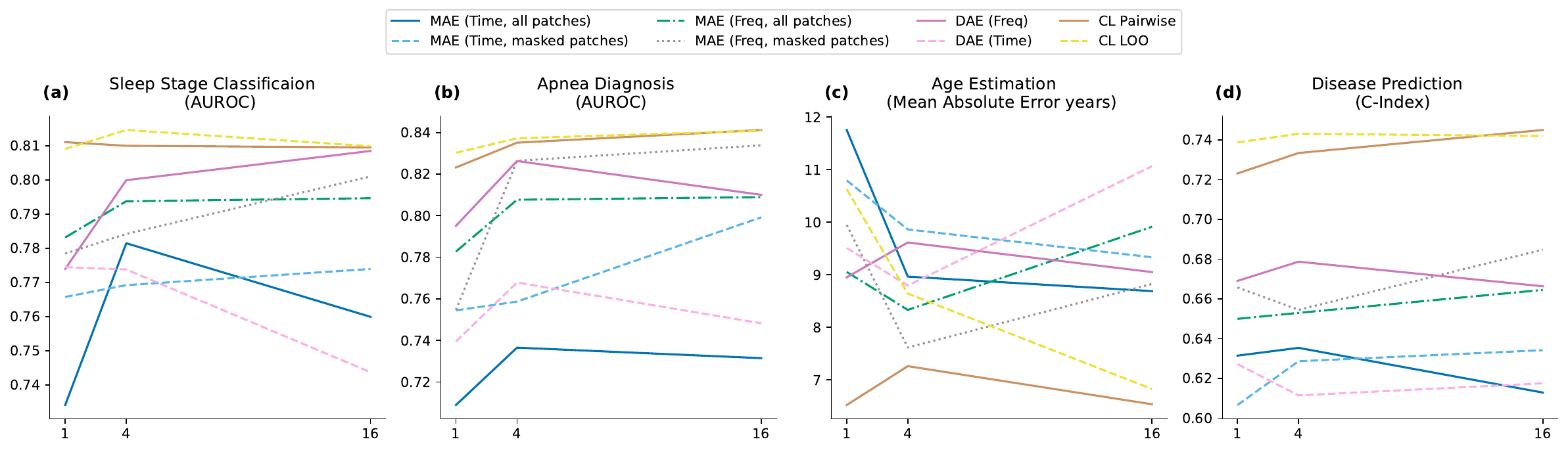}
    \caption{Analysis of self-supervised representation learning task convergence during pretraining across three epochs (1, 4, and 6). Training was conducted on a subset of the dataset containing 5,120 subjects to evaluate convergence speed on downstream tasks.}
    \label{fig:learning_curves}
\end{figure*}

We evaluated the downstream performance of SSRL methods in a compute-controlled setting using a fixed subset of 5,120 subjects, training for 1, 4, and 16 epochs. As shown in Supplementary \Cref{fig:learning_curves}, foundation models in this domain converge rapidly, with minimal performance gains beyond 4 epochs. In some cases, extended pretraining leads to decreased performance. The relative ranking of SSRL methods remains consistent with the best-performing models, with contrastive learning generally requiring less pretraining than other approaches.



\end{document}